# THE NEW APPROACH ON FUZZY DECISION TREES

Jooyeol Yun1, Jun won Seo, and Taeseon Yoon

1Department of Natural Science, Hankuk Academy of Foreign Studies, Yong-In, Republic of Korea jw987@naver.com, YunJooYeol98@gmail.com

Hankuk Academy of Foreign Studies, Yong-In, Republic of Korea tsyoon@hafs.hs.kr

## ABSTRACT

Decision trees have been widely used in machine learning. However, due to some reasons, data collecting in real world contains a fuzzy and uncertain form. The decision tree should be able to handle such fuzzy data. This paper presents a method to construct fuzzy decision tree. It proposes a fuzzy decision tree induction method in iris flower data set, obtaining the entropy from the distance between an average value and a particular value. It also presents an experiment result that shows the accuracy compared to former ID3.

## 1. INTRODUCTION

Decision trees have been widely used in data mining. They help us predict unseen tests in reasonable ways. As it eliminates the maximum amount of possibilities of committing errors using entropy, it is used for decision making and classification. Many algorithms such as ID3, C5.0 have been devised for decision tree construction. These algorithms are widely used in scientific test, image recognition, of sorts. But they have everything to do with handling certain data that can be firmly identified. They can't output data revealing the membership between more than two attributes. However, due to observation errors, uncertainty, and so on, data collecting in the real world contains a lot of fuzzy and uncertain data. There is a need for some algorithms which can be applied in real-world situation and also has a high accuracy. Data collecting in real world usually happens in fuzzy forms, which show degrees of membership. Using fuzzy forms, it is available to get output data that is member of many attributes in real-world situation. As a consequence, it is possible to obtain data in real world using fuzzy set theory in decision tree. Then fuzzy numbers are unable to construct decision trees because it is unable to calculate entropy. This paper is concerned with combining the fuzzy set and entropy for constructing fuzzy decision tree. Iris flower data set is going to be used. The average of many iris data can be obtained. The distance between the average of iris flower data and a specific iris data is going to be used as the index of entropy. It introduces a procedure of constructing fuzzy decision tree using Iris flower data set and the experiment that compares its performances with former ID3 algorithms.

## 2. RELATIVE RESEARCH

Decision trees are 'tree-like' frame that simulates decisions with a computer programming, and it is used in rule mining [3]. Decision trees consist of 3 nodes : Decision nodes, chance nodes, and end nodes. Decision nodes are commonly represented by squares, chance nodes are represented by circles, and end nodes are represented by triangles. Decision trees make some attributes end up into a leaf node from specific branches. Each attribute can be various, depending on individuals and groups. So decision trees calculate distribution of answers for attributes and place the attributes in an adequate tree form in each case. In other words, it determines the order of questions to be asked, so that it can eliminate the maximum amount of possibilities of committing errors. Entropy is used in ID3 algorithms, and information gain is used in C4.5 and C5.0.

Entropy was first quantified by Shannon[4] in 1948. Entropy stands for the degree of 'uncertainty'. Let set X the answers to attributes with each element being {x}. The formula of entropy H(X) can be defined as following equation. .

The function  stands for the possibility of each answer. When all the p(x) values are equal, the highest entropy is obtained, which implies that they are very uncertain data. In addition, higher entropy leads to more uncertain data. Information gain (IG) means classifying data by choosing an attribute. IG(A) for an attribute is expressed as   Sv stands for a subset of S when attribute A contains "v".

The formal function means the entropy of "S" and latter function means sum of entropy of subset Sv. Information gain(S,A) stands for the expected result of the decline of entropy when attribute A is used in distribution. Thus, the attribute with the highest information gain is the most discrimination attribute of the given set. Decision trees apply this conception to eliminate the most entropy and choose the order of attributes. Many algorithms are devised for decision tree construction by using this conception. Entropy is used in ID3 (Iterative dichotomiser 3), which was first proposed by Ross Quinian[2], and Information gain is used in C4.5, and C5.0. This is widely used in various situations, and it makes inapplicable in which needs a numerical decision. It is a method of structuring a decision tree in order of which information gain is large. Fuzzy sets are the ones whose elements have degrees of membership, which were introduced by Zadeh in 1965[7]. In the classical sets, a set is assessed in binary system which classifies an element into either "belonging(1)" or "not belonging(0)" to the set. On the contrary, a fuzzy set classifies an element using the membership function which is valued in the real unit interval [0,1]. It can be applied for uncertain data, which happens in real world situation. Showing it in function, fuzzy set μ(A) stands for the degree of membership for A. Then the membership function is defined as following. . In fuzzy decision tree, a fuzzy data(which means a degree) is given and it shows the extent to the data belongs to particular values.[6] Also, triangular membership function can be used in order to construct a decision tree that deals with uncertain data.[5] In order to calculate entropy, triangular membership function was used in a previous research. It made fuzzy attributes, which had membership in a few sets, certain attributes. Triangular membership function was used because of its simplicity, easy comprehension, and computational efficiency.

3. METHOD

The decision tree, which uses ID3, classifies data by using the conception of entropy. Due to many reasons, data collecting in real world contains fuzzy data. So the fuzzy set theory, which shows membership, needs to be applied to decision tree constructing. The method for constructing the fuzzy decision tree is turning the membership into the possibility in order to obtain entropy. Entropy can be defined as   . And in order to calculate entropy, all the possibilities should be summated to 1. However, fuzzy set does not always come to 1. The distance between the average of memberships and a particular value is going to be used in order to calculate entropy, which is the index of uncertainty.

Iris flower data set is going to be used as an example to explain. Iris flower data set is a multivariate data set that has been introduced by R.A Fisher as an example for discriminant analysis [1]. It consists of 4 features and 3 species of flower. Each flower has 4 features and can be described as fi[ai1, ai2, ai3, ai4]. 'a' stands for the features. It expresses a dot in a four dimensional space. In iris flower data set, many fi s are obtained up to the number of information. And each feature can be obtained as fuzzy number, and each feature can be assumed that it is divided into 3 parts(High, Medium, Low) and it has fuzzy membership value for each part. There would be a classification standard that determines fuzzy membership value for each part, as figure 1. For example, μa1 would represent 0.9 of μn and 0.1 of μm and 0 of μw. Then the membership values of 3 parts can be obtained in 4 features. As one flower has 3 features, 12 fuzzy membership values are obtained totally in one flower. It can be written as F=[μn(a1),μm(a1),μw(a1),……,μw(ai)]. This set 'F' can be expressed as a vector in a twelve dimensional space. It is information of one flower in iris flower data set. Also, the number of obtained F values is the number of information. The average of membership for each feature is obtained here by many F sets.

Decision trees need the value of entropy, which is the index of uncertainty, in order to eliminate the most possibilities of committing mistakes in the process of constructing it. Now, the average of information is obtained. So the distance between a particular value and the average value can be applied in the form of entropy, .

Figure 1

The coordinate of average is to be called Fi, and the coordinate of a particular value fk. Generally, the distance between two coordinate is defined as |Fi-Fk|. And the distance between these two values are going to be called "Zik"

The value of distance is going to be applied to the entropy. If the distance "Zik" becomes longer, the uncertainty becomes bigger. Else, if the distance "Zik" becomes shorter, the uncertainty becomes smaller. Consequently, the distance and the entropy are in inverse proportion. The distance has a value above 0, and the entropy is valued from 0 to 1. Therefore the entropy using the distance between two values can be defined as and it can be put into the ID3. This form shows that if Zik is getting closer to 0, which means a particular value is getting closer to the average value, the output number will lead closer to 1. This kind of form is suitable for ID3.

4. RESULT In the process of experiment, the data is divided into 2 parts for experimental convenience. And the data is the iris flower data set. The former ID3 classifies the data (length, width, and so on) by the aid of entropy. However, the new method that uses fuzzy membership function, determines the uncertainty with the distance from an average. Iris data has 4 features and each feature is divided into 2 parts. So the 4- dimensional data of each flower is expressed as the fuzzy 8-dimensional vector. The five fold cross- validation was used for experiment. In the five fold cross validation, a single model was retained as the validation, and the other four models were used as training data. 150 pieces of data were divided into 3 groups according to the classes. In two groups, data number 1 to 10 was used as test data, and the other was used as training data. The test data was changed regularly, like 1 to 10 and 11 to 20 and so on.

If the group 1 and the group 2 are compared, the result is obtained in this form.

Table 1: The results of ID3 in the five-fold method

| Exp. Count | Group A | Group B | Group C | Group D |
|---|---|---|---|---|
| 1 | 9 | 0 | 1 | 10 |
| 2 | 9 | 3 | 1 | 7 |
| 3 | 10 | 1 | 0 | 9 |
| 4 | 9 | 0 | 1 | 0 |
| 5 | | | | |

Table 2. The result of fuzzy decision tree in the five-fold method

| Exp. Count | Group A | Group B | Group C | Group D |
|---|---|---|---|---|
| 1 | 9 | 0 | 1 | 10 |
| 2 | 10 | 0 | 0 | 10 |
| 3 | 9 | 2 | 1 | 8 |
| 4 | 9 | 1 | 1 | 9 |
| 5 | | | | |

Group A: the number of cases, when the test determined group 1 from the real group 1

Group B: the number of cases, when the test determined group 1 from the real group 2

Group C: the number of cases, when the test determined group 2 from the real group 1

Group D: the number of cases, when the test determined group 2 from the real group 2

1 : The result of five- fold from the comparison between group 2 and group 3

2: The result of five- fold from the comparison between group 2 and group 3

3: The result of five- fold from the comparison between group 2 and group 3

4: The result of five- fold from the comparison between group 2 and group 3

Table 1 result showed 80 to 95 percent accuracy. Table 2 result showed it had 5 to 15 percent of committing errors. When fuzzy decision tree was compared with ID3, each five-fold result had different accuracy - fuzzy decision tree showed better accuracy in some cases, and ID3 showed better accuracy in the others. It means that new method of fuzzy decision tree show the similar performance with former ID3 algorithms. In the later research, the way for increasing accuracy will be studied.

5. CONCLUSION The paper is concerned with fuzzy sets and decision tree. We present a method of fuzzy decision tree that uses the distance from an average for the index of uncertainty. It proposes a fuzzy decision tree induction method for fuzzy data of which data is obtained by a membership function. An experiment is used in order to prove the validity, and to compare it with former ID algorithm. First, we applied fuzzy set theory because real world contains a lot of fuzzy numbers. Second we used iris flower data set to construct a decision tree. According to the result of the experiment, it showed similar accuracy compared to former ID3.

REFERENCES [1] Patrick, S.Hoey. (n.d.) 'Statistical Analysis of the Iris Flower Dataset', , (), pp. 1-5. [2] Quinlan, J. R. (1986). 'Induction of Decision Trees'. Mach. Learn. 1(1),pp. 81-106    [3] Ron Kohavi, Ross Quinlan (1999) 'Decision Tree Discovery', IN HANDBOOK OF DATA MINING AND KNOWLEDGE DISCOVERY, (), pp. 1-16. [4] Shannon, C.E. (1948) 'A Mathematical Theory of Communication', Bell System Technical Journal, (), pp. 377-423 & 623-656. [5] TIEN-CHIN WANG, HSIEN-DA LEE (2006) 'Constructing a Fuzzy Decision Tree by Integrating Fuzzy', ACOS'06 Proceedings of the 5th WSEAS International Conference on Applied Computer Science , (), pp. 306-311 [6] Yuan, Y., Shaw, M.J (1995) 'Induction of fuzzy decision trees', Fuzzy Sets and Systems, 69(2), pp. 125-139 [7] Zadeh, L.A. (1965) 'Fuzzy Sets', (), pp. 1-16.

Authors

Short Biography

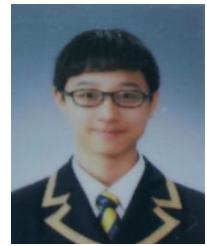

**Jooyeol Yun** was born in BunDang, Korean, in 1998. Since 2014, he has been studying in the Department of Natural Science, Hankuk Academy of Foreign Studies. He has an interest in Artificial Intelligence.

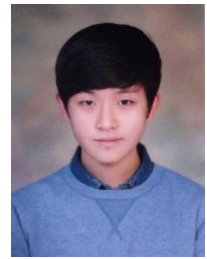

**Jun Won Seo** was born in Seoul, Korea, in 1998. Since 2014, he has been studying in the Department of Natural Science, Hankuk Academy of Foreign Studies. He has an interest in Artificial Intelligence.

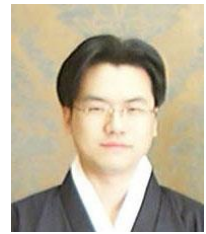

**Taeseon Yoon** was born in Seoul, Korea, in 1972. He was Ph.D. Candidate degree in Computer education from the Korea University, Seoul, Korea, in 2003. From 1998 to 2003, he was with EJB analyst and SCJP. From 2003 to 2004, he joined the Department of Computer Education, University of Korea, as a Lecturer and Ansan University, as a Adjunct professor. Since December 2004, he has been with the Hankuk Academy of Foreign Studies, where he was a Computer Science and Statistics Teacher. He was the recipient of the Best Teacher Award of the Science Conference, Gyeonggi-do, Korea, 2013.